# A NOVEL METHOD FOR SPEECH SEGMENTATION BASED ON SPEAKERS' CHARACTERISTICS

Behrouz Abdolali [1] and Hossein Sameti [2]

[1] Faculty and Research Center of Communication and Information Technology.
IHU Tehran, Iran
`abdolali@ce.sharif.edu`
[2] Department of Computer Engineering, Sharif University of Technology
`sameti@sharif.edu`

## ABSTRACT

*Speech Segmentation is the process change point detection for partitioning an input audio stream into regions each of which corresponds to only one audio source or one speaker. One application of this system is in Speaker Diarization systems. There are several methods for speaker segmentation; however, most of the Speaker Diarization Systems use BIC-based Segmentation methods. The main goal of this paper is to propose a new method for speaker segmentation with higher speed than the current methods - e.g. BIC - and acceptable accuracy. Our proposed method is based on the pitch frequency of the speech. The accuracy of this method is similar to the accuracy of common speaker segmentation methods. However, its computation cost is much less than theirs. We show that our method is about 2.4 times faster than the BIC-based method, while the average accuracy of pitch-based method is slightly higher than that of the BIC-based method.*

## 1. INTRODUCTION

The process of speaker change detection from one speaker to another is an important task in many speech processing applications. This task is done before audio indexing, speaker identification, automatic transcription, information extraction, speech summarization and retrieval [1][2][3]. An audio stream can be segmented into various homogeneous parts by recognizing the specific speech characteristics of individual speakers. This process is commonly known as speaker change detection or speaker segmentation. In recent years, there are three major categories of audio segmentation techniques: metric-based, model-based and hybrid methods. Each one has its advantages and disadvantages which will be discussed in the next section.

The most common speaker segmentation methods are those metric-based ones which are used Bayesian Information Criterion (BIC). Since these methods suffer from a great amount of computations, they are very time consuming; however, these methods are highly accurate. Achieving a method which has acceptable accuracy along with high computation speed is very desirable and useful for real time systems. In this paper we will discuss about a proposed method for speech segmentation that doesn't need previous information about speakers and also hasn't heavy computation. In other word, we want to solve the problem of low speed of metric-based segmentation methods. For solving this problem we have proposed to use of speaker's pitch frequency information. In this method the change points are detected according to speakers' pitch frequency function. Of course using this function has some problems such as increasing error, but by using some techniques we decrease the errors.

## 2. A SURVEY ON SPEAKER SEGMENTATION METHODS

Various speaker segmentation algorithms have been proposed. These algorithms can be categorized into the following categories: metric-based, model-based and hybrid segmentation algorithms.

Another approach called decoder-based has also proposed. It is assumed that the sentences uttered by different speakers in a conversation are delimited by pauses [4]. As a consequence the segmentation relies on the accuracy of an inter speaker silence detector which usually works by measuring the energy or zero crossing rate of each segment and comparing it to a predefined or adaptively estimated threshold. The main drawback of this approach is no direct connection exists between a detected silence and an actual speaker change. Because of this assumption, this method isn't used for actual meeting.

In metric-based methods, first an acoustic distance criterion has been defined and then two adjacent windows are shifted along the audio stream. Depending on the application the analysis window may overlap or not. By measuring the distance between these two windows the similarity between these segments is evaluated and a distance curve is formed. This distance curve was often low-pass filtered and the locations of peaks were chosen to be acoustic changing points by heuristic thresholds. Most of the distance measure criterions come from the statistical modeling framework. The feature vectors in each of the two adjacent windows are assumed to follow some probability density(usually Gaussian) and the distance is represented by the dissimilarity of these two densities, e.g., the Kullback-Leibler distance(KL,KL2)[5], generalized likelihood ratio(GLR)[6] and Bayesian Information Criterion(BIC) [7][8][9].

The metric-based methods have the advantage of not requiring any prior knowledge on the number of speakers, their identities, or signal characteristics; but they have some disadvantages: (1) it is difficult to decide an appropriate threshold. (2)Each acoustic changing point is detected only by its neighboring acoustic information. (3) To deal with homogenous segments of various lengths, the length of window is usually short (typically 2 seconds), so the feature vectors could be insufficient to obtain robust distance statistics.

In the model based approach, a set of models is derived and trained for different speaker classes from a training corpus. It assumes that a speaker change is likely to occur at the time indexes where the model's identification decision changes from one speaker to another. As a result, prior knowledge is a prerequisite to initialize the speaker models. The models can be created by means of hidden Markov models (HMMs) [10],[11] ,[12] ,Gaussian mixture models (GMM) [13], [14] or support vector machines (SVM) [15], [16], [17].

Hybrid based methods combine metric and model based techniques [18]. A set of speaker models are created by presegmenting the input audio signal using metric based approaches. Then the model based segmentation is applied to yield a more refined segmentation. In [19], HMMs are combined with BIC. Another hybrid system is introduced in [20] where two systems are combined namely LIA system, which is based on HMMs and the CLIPS system, which performs BIC based speaker segmentation followed by hierarchical clustering.

## 2.1. Speaker Segmentation based on BIC measure

Since BIC-based methods are the most common segmentation methods used today, we focus on these methods in detail. BIC is a criterion for choosing a model for a group of data which is proposed by Schwarz[21]. Suppose we have a group of data (X) and a Model for describing these data (M). The BIC criterion for this model is shown in Equation 1.

$$BIC_M = \log p(X|M) - \lambda \frac{\#M}{2} \log N \quad (1)$$

In this equation, P(X|M) is the likelihood value of data X to model M. #M represents the number of free parameters in Model M. N represents the number of samples in data X. In the other word, BIC measures the likelihood of the model and data and scores the model[22]. In the above equation, λ is the penalty factor. If λ is set to zero, BIC changes to GLR[23]. To achieve the expected performance for a specific corpus, we could adjust λ value[24]. As described in [21], maximizing BIC results in maximizing the expected value of likelihood of model and data. Therefore, BIC could be used to select the best model of a group of data[21,22].

Suppose X= $x_i \in R^d$, i = 1,2,...,N is a sequence of feature vectors of d dimensions which are extracted from a speech frame. In such applications, usually MFCC feature vectors are used. BIC criteria don't have any pre knowledge about the type of feature vectors. Therefore, this criterion could be used even when other feature vectors are used. We suppose that in a frame there are at most two speakers (one speaker boundary). Therefore, the problem of checking if a single speaker change point exists in the frame, could be transformed to a model selection problem[24]. To do this, we consider two adjacent windows (X and Y) around hypothetical time instance (b). Our objective is to decide if a speaker change occurs in this instance time or not.[18].

- Model M1 supposes that all samples in X are independent and evenly distributed by a multivariate Gaussian process.

$$M_1 : \quad Z = z_1, z_2, \ldots, z_N \sim N(\mu_z, \textstyle\sum_z) \quad (2)$$

- Model M2 suppose that X is created by two multivariate Gaussian processes. One from the beginning of the frame to time b, and one from time b to the end of the frame.

$$M_2 : \quad Z = X + Y \quad (3)$$

$$X = z_1, z_2, \ldots, z_b \sim N(\mu_x, \textstyle\sum_x) \quad (4)$$

$$Y = z_{b+1}, z_{b+2}, \ldots, z_N \sim N(\mu_y, \textstyle\sum_y) \quad (5)$$

These two hypotheses are shown in Figure 1.

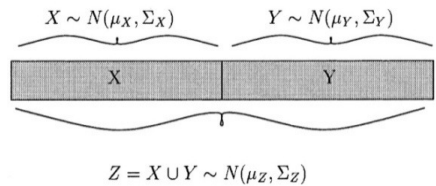

Figure 1. Hypothetic models for segmentation of one speech frame[25].

The difference between BIC scores of the models is expresses as $\Delta BIC$ as shown in Equation 6.

$$\Delta BIC\{x,y\} = BIC(M_2, Z) - BIC(M_1, Z)$$

$$= log\, p(X|\,\widehat{\mu_x}, \widehat{\Sigma_x}) + log\, p(Y|\,\widehat{\mu_y}, \widehat{\Sigma_y}) - log\, p(Z|\,\widehat{\mu_z}, \widehat{\Sigma_z}) - \frac{1}{2}\lambda\left(d + \frac{1}{2}d(d+1)\right)log\, n$$

$$= \frac{n}{2}log|\widehat{\Sigma_z}| - \frac{n_x}{2}log|\widehat{\Sigma_x}| - \frac{n_y}{2}log|\widehat{\Sigma_y}| - \frac{1}{2}\lambda\left(d + \frac{1}{2}d(d+1)\right)log\, n \qquad (6)$$

In Equation 6, $\widehat{\Sigma_1}, \widehat{\Sigma_2}, \widehat{\Sigma}$ are estimations of covariance matrices of corresponding data with maximum correctness. Operator | . | is determinant operator and $d$ is the dimension of cepstral feature vector. In this equation, $\lambda$ is called penalty factor, when it is set to zero, BIC changes to GLR[23].

If $\Delta BIC = BIC(M2) - BIC(M1) > 0$, it means that the score of describing data by two Gaussian distributions (M2) is more than that of describing data by a single Gaussian distribution (M1). Therefore, data is not uniform and there is a speaker change point.

We should notice that ΔBIC is used to detect only a single acoustic change point in the speech stream. As a result, it is necessary to use other algorithms to detect more change points. For this purpose, underline{sequential detecting algorithms} were proposed in [23]. We can rewrite $\Delta BIC$ as a function of change point (b). If the number of feature vectors in X or $n_x$ is equal to b and $n_y$ is equal to n-b, $\Delta BIC$ will be as shown in Equation 7.

$$\Delta BIC_b\{x,y\} = \frac{n}{2}log|\widehat{\Sigma_z}| - \frac{b}{2}log|\widehat{\Sigma_x}| - \frac{n-b}{2}log|\widehat{\Sigma_y}| - \frac{1}{2}\lambda\left(d + \frac{1}{2}d(d+1)\right)log\, n \qquad (7)$$

Based on $\Delta BIC$ value, segmentation of speech stream into two parts is correct when $\Delta BIC(b) > 0$. Positive value means Model M2 best describe the signal and change point of b exists. In the following sections, some common algorithms for finding change points are explained.

### 2.1.1. Increasing size window method for calculating $\Delta BIC$

This method is used to detect multiple change points in a stream. As shown in Figure 2, we consider an initial size for the window in which $N_{ini}$ feature vectors exist. This window continually increases its size by $N_g$ until a change point is found based on BIC criterion. A higher band for the window size, $N_{max}$, is also determined. If a change point is detected before reaching the window size to $N_{max}$, the point is marked and the process begins from this point with initial window size. Otherwise, after reaching $N_{max}$, the window is shifted by $N_s$ samples and the process repeats[23]. It is notable that, as the time progresses and window size increases more processing is needed. Therefore, this method suffers lower performance and requires more processing power.

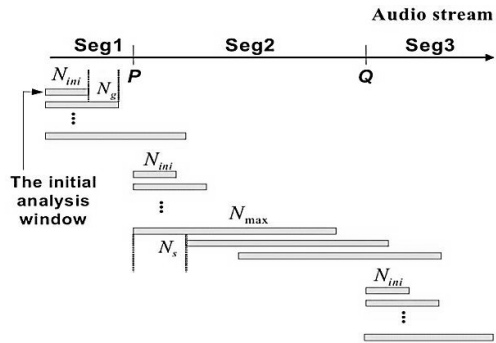

Figure 2. Increasing size window algorithm for speaker change detection[24].

### 2.1.2. Fixed-Size sliding window method for calculating $\Delta BIC$

In this method, one window with fixed-size is considered and by sliding it across the steam, $\Delta BIC$ is calculated. Window size depends on the length of the stream. Setting the window size and λ parameters to optimum values results in higher performance and accuracy. It is obvious that this method needs less processing power. Nevertheless, its accuracy is lower than previous one. According to experiment, for short streams about some of minutes, the window size of 1 second is well[23].

In this paper, we implemented the first method to compare the best accuracy achieved by BIC-based methods with our proposed method.

## 3. THEORETICAL TOPICS ABOUT PITCH FREQUENCY

In this section, after a short looking at pitch frequency characteristics, we review some important pitch extraction methods.

### 3.1. Pitch Frequency and Its Characteristics

Pitch frequency is the fundamental harmonic of the speech signal. In the other word, it is the fundamental frequency of the human vocal cords' vibration. More like the sine wave is the wave form of the signal, more clearer the sense of frequency and less clearer the sense of pitch.[26] Likewise, more harmonic to each other are frequency components, more clearly the sense of pitch and less clearer the sense of frequency[27,28,29]

### 3.2. Pitch frequency extraction methods

There are several methods used for the estimation of fundamental frequency of $f_0$. Nonetheless, it is difficult to propose a method which estimates $f_0$ well, without considering the content of the signal. Therefore, in the environment in which both music and speech signals exist, pitch estimators should be accurate in both fields. The difficulty of detecting $f_0$ in a waveform depends on the wave form. It means that if the waveform contains less high harmonics in the frequency spectrum, or the power of higher harmonics is low, $f_0$ will be simpler to detect[30].

Pitch frequency determination algorithms called PDA, are of a great importance in many speech processing algorithms. In the following sections, pitch determination methods based on autocorrelation function, cepstrum method, linear prediction coding (LPD) method, and average magnitude difference (AMDF) function will be discussed.

### 3.2.1. Pitch Detection using autocorrelation function (ACF)

Our perception of pitch frequency is strongly related to our perception of waveform periodicity in the time domain. The method which can determine the fundamental frequency of a signal based on its waveform is the autocorrelation method[31].autocorrelation function of a signal s[n], is shown in Equation 8 in which τ is the delay or time shift. Calculating this function and detecting its maximal points, we can estimate pitch frequency of signal s[n].

$$R(\tau) = \sum_{n=0}^{N-1} s(n).s(n+\tau) \quad (8)$$

### 3.2.2. Pitch Detection using cepstral method

Cepstral analysis provides a method for the pitch estimation. Suppose that a sequence of speech samples is the result of applying convolution function on the sequence of glottal excitation e[n] and vocal tract's discrete impulse response θ[n]. In the frequency domain, convolution operator changes to multiplication operator. Using the characteristics of Algorithm function (log(A.B)=log(A)+log(B)), multiplication operator could be changed to addition operator. Finally, the real cepstrum of a signal expresses by the formula s[n]=e[n]*θ[n], is c[n] which is shown in Equation 9.

$$c[n] = \frac{1}{2\pi} \int_{-\pi}^{\pi} \log |S(\omega)| e^{jn\omega} d\omega \qquad (9)$$

In the above formula S(ω) is:

$$S(\omega) = \sum_{n=-\infty}^{\infty} s[n] e^{-jn\omega} \qquad (10)$$

Therefore, cepstrum is the result of applying Fourier transform on the logarithm of amplitude of the signal spectrum. If the logarithm of amplitude of the signal spectrum contains several harmonics which are placed at regular distance from each other, Fourier transform of the spectrum includes a peak which corresponds to the distance between harmonics. This peak, in fact, is the fundamental frequency or pitch of the signal.

### 3.2.3. Pitch Detection using average magnitude difference function (AMDF)

AMDF concept is very close to ACF concept, except that in this function amplitude difference between the frame and its delayed version is estimated instead of estimating likeness between them. AMDF calculation is shown in Equation 11. In this equation, τ is the time range in terms of speech samples. The value of τ, for which AMDF(τ) in a specific range is minimum, is chosen as the period of the pitch.

$$AMDF(\tau) = \sum_{i=0}^{n-1} |S(i) - S(i + \tau)| \qquad (11)$$

In the other word, delayed version of the frame is moved n times and the absolute value of the summation of difference in overlapping sections is calculated to produce n AMDF value. Pitch value is the result of division of sampling frequency by speech sample corresponds to the first local minimum in AMDF function.

## 4. SPEAKER SEGMENTATION USING PROPOSED METHOD

Distance-based speaker segmentation methods, such as BIC method, use cepstral features like Mel-Frequency Cepstral Coefficient (MFCC). However, there are other feature vectors which can be used for this purpose. In addition, prosodic features like pitch frequency can be used to facilitate distinguish between voice and silence. Pitch frequency changes diagram is a well-suited means for speaker change detection[32]. There are three reasons to use pitch frequency for speaker change detection:

1- Every speaker has its own pitch frequency which differs from others.

2- When the speaker changes, pitch frequency diagram has rapid changes.

3- For speech segments less than one second, other methods such as BIC-based methods which utilize MFCC features, suffer lack of speaker change detection accuracy. Since, there is not enough information in such a short segments to calculate meaningful MFCC vectors.

To make it possible to use pitch frequency information for speaker segmentation, the preliminary stage is to extract pitch frequency using one of the above mentioned methods. We choose AMDF method because of its calculation speed.

After calculation of pitch frequency value for each individual speech frame, we should use this information for speaker change detection. Since rapid changes in the pitch frequency diagram can be used to indicate speaker change, we use this indication afterwards. Suppose to have divide speech stream to N windows and extract pitch frequency for each window independently. To analyze pitch frequency changes, we use derivation function which is shown in Equation 12.

$$Diff(x) = |pitch(n + 1) - pitch(n)| \qquad (12)$$

Then we should determine a threshold value by which we evaluate pitch frequency changes. If derivative function of pitch frequency at one point is above this defined threshold, we consider that point as speaker change point. We define this threshold as 0.7 of maximum difference between higher and lower pitch frequency in the stream. After extracting speaker change points, we could calculate beginning and end of speech segments. This method has very high speed. Nonetheless, its detection accuracy compared with BIC-based methods is less. In the following section accuracy improvement method will be explained.

## 5. ACCURACY IMPROVEMENT OF PROPOSED METHOD

Detailed looking at pitch frequency changes diagram, we noticed that its variation is very sharp. Even during the speech of one speaker, it is be possible to have rapid pitch frequency changes. In these cases, False Alarm (FA) rate error increases. This is the result of considering every point above the threshold as speaker change point.

Another problem with this method is the possibility of speaker change while pitch frequency changes is not very quick to be above the predefined threshold. Therefore, we may miss these points of speaker changes and Miss Detection (MD) rate error may be increased. To solve this problem one way may be to choose the threshold value lower to place below the missed points. Obviously, this is not an efficient way, because using the lower threshold, many other points which are not true speaker change points will be placed above the threshold and incorrectly will be reported as change points. As a result, MD rate error decreases at the expense of increase in FA rate error. Using this method decrease the accuracy of speaker change detection method.

To cope with this problem, we should apply a function on pitch frequency change function which increase small changes and preserve large changes. Gamma correction function best suited for this purpose.

$$Gamma\big(f(x)\big) = C.f(x)^{\gamma} \qquad (13)$$

We apply this function on pitch frequency change diagram. Considering Figure 3, and above mentioned problem, it is clear that for our application, we should use γ<1 which based on our experiments, its best value is 0.3.

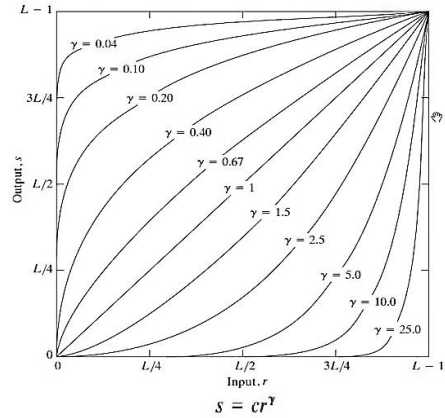

Figure 3. Gamma Correction function[33]

Applying this function, MD rate error decreases. However, since there were false rapid changes before applying gamma correction function, the problem of high FA rate error is remained. To reduce FA rate error, we could benefit the idea used in BIC segmentation. It means that we consider all points above the threshold as speaker change points and investigate correctness of change detection using a small BIC window of length 0.1s.experimental results show that applying this method, results in egregious improvement in detection accuracy. In Figure 4, flowchart of proposed change detection algorithm is shown.

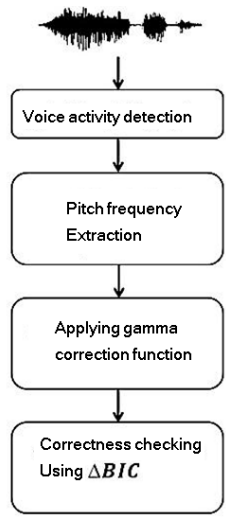

Figure 6. Segmentation flowchart for proposed RPSS method

## 6. THE EVALUATION MEASURES FOR SEGMENTATION METHODS

It is obvious that for each system, standard evaluation method should be introduced. It is also necessary for segmentation systems. In this section we discuss about these measures and standard evaluation methods for segmentation systems. Results of this paper are based on these methods.

Operation of the system could be analyzed for recorded sessions. However, valid results are those which are based on sessions in a speech corpus. There are several corpuses some of the important ones are NIST, AMI and TMIT.

For evaluation of segmentation systems, some measures are used which are the comparison between detected change points and real change points in the corpus under investigation. The most important measures used, are %FD and %FR which are calculated as shown in Equations 14,15.

$$\%FD = \frac{\# \ false\_detections}{total\_amount\_of\_detections} \qquad (14)$$

$$\%FR = \frac{\# \ missed\_detections}{total\_amount\_of\_true\_change\_points} \qquad (15)$$

false_detections: number of points which are not real change points in the reference corpus; but, are detected by the system as change points. These points are called False Alarm (FA).

total_amount_of_detections: total number of points detected by the system as change points.

missed_detections: number of points which are real change points in the reference corpus; But, are not detected by the system as change points. These points are called Missed Detection (MD).

total_amount_of_true_change_points: total number of points correctly detected by the system as change points.

To determine the accuracy of segmentation method, F measure is defined as shown in Equation 16.

$$F = \frac{2 * (1 - FD) * (1 - FR)}{2 - FD - FR} \qquad (16)$$

## 7. EXPERIMENTAL RESULTS

In this section using diagrams resulted from changing important parameters involved in the calculations of improved proposed method, we want to investigate the effect of these parameters on the accuracy and performance of the proposed method. In this paper we apply our method on four of AMI sessions. These sessions are selected randomly. The names of the sessions are included in corresponding tables. Diagrams show the average values achieved in four experiments.

Threshold coefficient is a parameter which determines the acceptable value of pitch changes compared with global maximum. This means that if this factor is 0.7, changes above the 0.7 of global maximum are considered as speaker change points. It is important to consider the effect of changes in this parameter on the accuracy and performance of the proposed method. Higher value of this parameter, results in less FA and more MD and vice versa. Experimental results show that the value of 0.75 is well suited for our purpose.

Based on the diagrams, we understand that increase in the value of γ for gamma correction, has a dramatic effect on the accuracy of the method. The best value for γ is 0.3 based on the results.

Table 1 shows the value of parameters involved in the accuracy and performance of the proposed method for each AMI session to achieve the best accuracy.

After examining all parameters, we achieve the optimum point for the accuracy and performance. Now we should compare our method with implemented BIC-based method. Table 2 summarizes results of two methods in their optimum point in terms of accuracy. Based on data in Table 2, we could conclude that segmentation using proposed method which is based on changes in the pitch frequency, has the advantage of improving the performance by 2.4 times and increasing the accuracy by 1% .

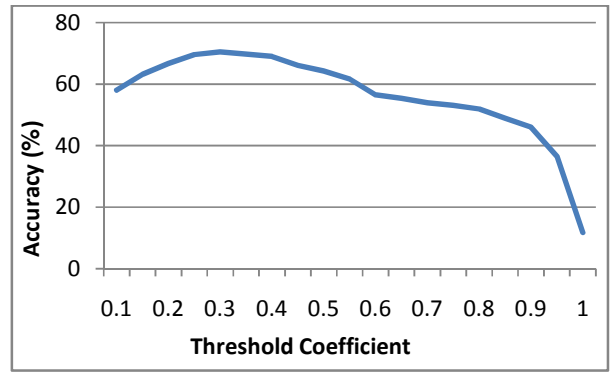

Figure 5. The effect of threshold coefficient on the accuracy of proposed method

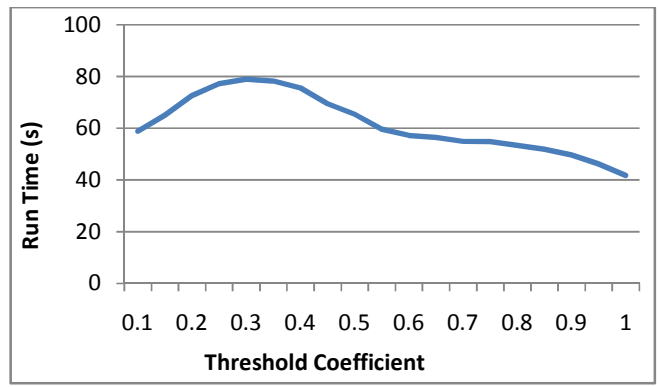

Figure 6. The effect of threshold coefficient on run time of proposed method

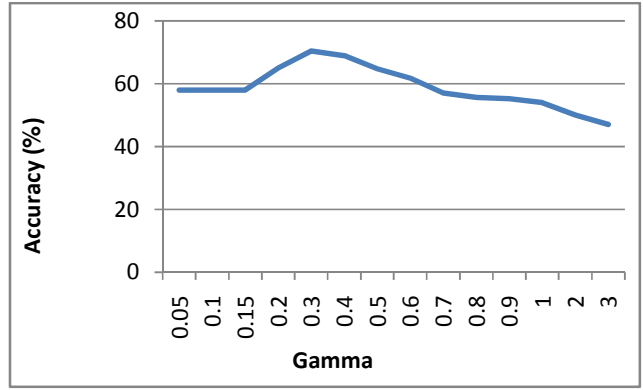

Figure 7. The effect of Gamma value on the accuracy of proposed method

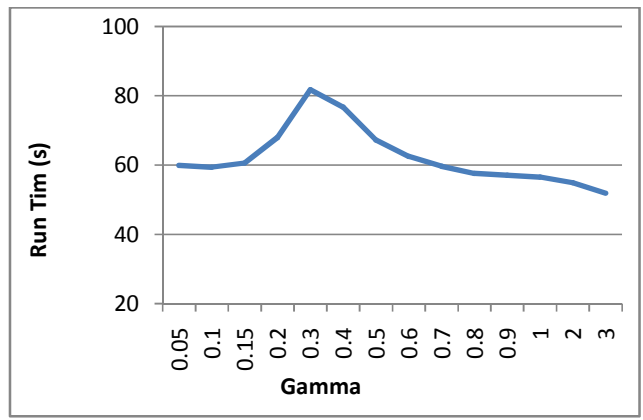

Figure 8. The effect of Gamma value on run time of proposed method

Table 1. The comparison between accuracy and performance of proposed method and BIC-based method with increasing size window

| File name (*.WAV) | Pitch Thresh Coef. | Mfcc Win Size (Sample) | Mfcc Overlap Size | $\lambda$ | BIC Win for Pitch (sec) | $\gamma$ | %FD | %FR | %F | Run Time (Sec) |
|---|---|---|---|---|---|---|---|---|---|---|
| ES2002a_p1 | 0.325 | 200 | 120 | 1 | 0.4 | 0.3 | 42.07 | 1.26 | 73.02 | 87.86 |
| ES2002b_p2 | 0.325 | 200 | 120 | 1.2 | 0.2 | 0.3 | 41.02 | 5.61 | 74.04 | 86.57 |
| ES2002c_p1 | 0.325 | 200 | 120 | 1.1 | 1.6 | 0.25 | 43.93 | 11.59 | 71.5 | 134.93 |
| ES2002d_p1 | 0.325 | 200 | 120 | 1 | 0.2 | 0.3 | 37.97 | 1.29 | 76.56 | 80.19 |

Table 2. The best values for parameters involved in the proposed method

| File name (*.WAV) | Length (Sec) | BIC Based Segmentation | | | | Pitch Based Segmentation (Proposed Method) | | | | Speed Up |
|---|---|---|---|---|---|---|---|---|---|---|
| | | %FD | %FR | %F | Run Time (Sec) | %FD | %FR | %F | Run Time (Sec) | |
| ES2002a_p1 | 365 | 42.87 | 0.63 | 72.55 | 328.98 | 42.07 | 1.26 | 73.02 | 87.86 | 3.74 |
| ES2002b_p2 | 390 | 41.62 | 5.62 | 73.56 | 190.85 | 41.02 | 5.61 | 74.04 | 86.57 | 2.2 |
| ES2002c_p1 | 407 | 44.07 | 1.45 | 71.36 | 195.52 | 43.93 | 11.59 | 71.5 | 134.93 | 1.44 |
| ES2002d_p1 | 360 | 38.88 | 0 | 75.87 | 176.38 | 42.07 | 1.26 | 73.02 | 87.86 | 2.19 |

## 8. CONCLUSION

In this paper a rapid and accurate method of speaker segmentation for speaker diarization applications is proposed. This method is based on pitch frequency changes and is better than BIC-based method in terms of run time. Also its disadvantage of lower accuracy is amended using innovative techniques namely Gamma correction function and BIC-based double checking of candidate points. Taking advantage of the novel techniques, we achieve a performance of 2.4 times faster than the BIC-based methods while benefiting the same accuracy.